\documentclass[a4paper,fleqn]{cas-sc}

\usepackage[authoryear,longnamesfirst]{natbib}
\usepackage{graphicx}
\graphicspath{ {./doc/images/} }
\usepackage{placeins}
\newcommand\MyBox[2]{
  \fbox{\lower0.75cm
    \vbox to 1.7cm{\vfil
      \hbox to 1.7cm{\hfil\parbox{1.4cm}{#1\\#2}\hfil}
      \vfil}%
  }%
}
\usepackage[nopar]{lipsum}
\usepackage{array}
\usepackage{multirow}
\usepackage{amsmath}
\usepackage{hyperref}

\makeatletter

\begin{document}
\let\WriteBookmarks\relax
\def\floatpagepagefraction{1}
\def\textpagefraction{.001}
\shorttitle{Stacked Generalizations in Imbalanced Fraud Data Sets using Resampling Methods}
\shortauthors{Kerwin and Bastian}

\title [mode = title]{Stacked Generalizations in Imbalanced Fraud Data Sets using Resampling Methods}                      
\author[1]{Kathleen Kerwin}[orcid=0000-0002-8065-2628]
\ead{kathleenkerwin2016@u.northwestern.edu}
\address[1]{Department of Data Science,
          Northwestern University,
          Evanston, Illinois 60201}

\author[2]{Nathaniel D. Bastian}[orcid=0000-0001-9957-2778]
\ead{ndbastian@northwestern.edu }
\address[2]{Department of Data Science,
          Northwestern University,
          Evanston, Illinois 60201}

\begin{abstract}
This study uses stacked generalization, which is a two-step process of combining machine learning methods, called meta or super learners, for improving the performance of algorithms in step one (by minimizing the error rate of each individual algorithm to reduce its bias in the learning set) and then in step two inputting the results into the meta learner with its stacked blended output (demonstrating improved performance with the weakest algorithms learning better).  The method is essentially an enhanced cross-validation strategy.  Although the process uses great computational resources, the resulting performance metrics on resampled fraud data show that increased system cost can be justified.  A fundamental key to fraud data is that it is inherently not systematic and, as of yet, the optimal resampling methodology has not been identified.  Building a test harness that accounts for all permutations of algorithm sample set pairs demonstrates that the complex, intrinsic data structures are all thoroughly tested.  Using a comparative analysis on fraud data that applies stacked generalizations provides useful insight needed to find the optimal mathematical formula to be used for imbalanced fraud data sets.
\end{abstract}


\maketitle

\section{Introduction}\label{Introduction}

Predicting fraud is challenging due to inherent issues in fraud data structure since the crimes are committed through trickery or deceit with an ever-present moving target of changing modus operandi to circumvent human and system controls. In addition to these complex and unsystematic characteristics, there is extreme imbalance in the data due to the class distributions between fraudulent and normal transactions. Rather than testing across data sets, the study focus is on how data and algorithm level techniques perform in combination, as well as how to select the best performing machine learning model.  Stacked generalizers provide an additional layer of generalization error correction, improving upon using a single classifier which is helpful because even the smallest improvement in fraud detection can have significant and far reaching ramifications.  Designing the machine learning system includes combining data and algorithm level development using the stacked generalization architecture.  The purpose is to test baseline and stacked classifier pairs with meta learners and combine their outcomes to compare which machine learning models have the optimal performance. Meta learners of stacks outperform baseline learners but are not statistically significant, although the improvement of the total cost of losses in the transactions in production may be substantial.  

According to the Association of Certified Fraud Examiners (ACFE), businesses worldwide lose 5\% of their revenues annually to fraud from internal and external sources \citep{ACFE2018}. "Fraud consists of some deceitful practice or willful device, resorted to with intent to deprive another of his right, or in some manner to do him an injury".\footnote{See the Law Dictionary https://thelawdictionary.org/fraud/}    The United Nations Office of Drugs and Crime (UNODC) estimates that approximately 2--5\% of global Gross Domestic Product, or \$800 billion--\$2 trillion in current US dollars is money laundered globally based on predicate crimes that include trafficking in drugs, psychotropics, stolen goods, arms, humans, racketeering, terrorism, terrorism financing, sexual exploitation, counterfeiting of currency or goods, corruption, bribery, fraud, environmental crimes, murder, kidnapping, theft, extortion, forgery, piracy, smuggling, insider trading, market manipulations, and tax crimes, for example \citep{UNODC2019}. 

Machine learning does not find fraud.  Machine learning algorithms learn using known fraud cases and then make inferences on new data finding similar configurations that may be fraudulent transactions.  These results are then sent to fraud investigators for confirmation, which is a labor-intensive endeavor.  Due to the extraordinarily high volume and complexity of transactions that take place daily in business, machine learning has become an invaluable tool used to identify potential fraud cases. Fraudulent transactions are rare cases of interest marked as the positive minority class with the normal transactions categorized as the negative majority class. The skewed data set requires special handling for the class imbalance, which favors the majority class often ignoring the minority class due to the large disparity in their relative numbers; this leads to inaccurate metrics and results. The two-class problem requires supervised machine learning using binary classification modeling.

Identification of illegal activity may be made through system monitoring for thresholds or by using filters, software matching fraud typologies, anomaly detection, or human detection by a customer complaint, law enforcement or a tip from a customer facing employee among other sources.  One of the major challenges is that fraudulent transactions may appear as ordinary, normal transactions but are unlawfully committed through trickery or deceit. For example, identity theft may appear as a genuine business operation with the fraud pattern identified from a source not within the electronic event.  The transaction may be identified as fraudulent by law enforcement detective work with the fraud pattern not entirely appearing within the electronic business data or operation. If the fraud pattern information is not added into the electronic event flagged as fraudulent in the data set, it will predict similar non-fraudulent examples as true positive transactions (which is the definition of a false positive).  Compare this example to cancer detection where the actual lab test observations can be included in the data set with the indications of cancer inherently existing in the data.  For machine learning to identify similar patterns of fraud, additional relevant information may need to be reintroduced into the feature attributes or values. 

An added complication for fraud patterns is the possibility of a lack of consistent structure since criminals are working to obscure their identity, attempting to get away with the crime which uses a "throw everything at the wall approach" until something works while making the paper trail as complex as possible to unravel.  Contrast this with fields in which data set observations in math, science, or technology have systematic patterns with predictable organization. Although ever changing, regular and repeated order may be found in the fraud typology\footnote{Typology is defined as classification according to general type.} and possible electronic systems paths to accomplish the crime before controls limit access.  

Fraud is becoming increasingly profitable and being committed by organized crime where the exploitation of discovered weaknesses in the financial system are taken advantage of until controls are put in place \citep{Mills2017}.  Criminals then search for other means to exploit businesses, which means an ever-moving target for investigators to prevent and detect. Crime rings study how financial institutions operate and work around controls by fragmenting transactions into discrete units between organizations to obfuscate complex criminal collaborative efforts \citep{Brothers2015b}. The failure of the ability to see the more comprehensive picture of the illegal activity results in institution losses due to these crimes being written off or written down, as well as the perpetrators being able to disguise the crime typologies and identity of the criminals. Legal barriers in sharing information across financial institutions and law enforcement agencies (unless regulations trigger 314 (b) of the US Patriot Act based on suspicious activity or evidence of a crime) further encourage crime rings to exploit the inability of coordinated tracking and investigating \citep{Brothers2015a}. The Clearinghouse paper on "Reforms to the AML/CFT framework for national security and law enforcement" \citep{ClearingHouse1853} recommended illuminating and enhancing the range of information sharing.

Looking from the outside in, criminals try any and everything to defraud businesses and individuals from their assets.  From the inside out, criminals pay insiders for their knowledge, expertise, and assistance in identifying ways to find loopholes and circumvent manual and system controls. The only limitation for criminals in committing fraud is their creativity and ingenuity. Criminals continually exploit the electronic financial system to defraud consumers and businesses by finding weaknesses in the system including in audit controls.  When these weaknesses are found and fixes applied, criminals simply update their strategy and find other means of committing crimes. Since these fraud events are continually changing when discovered, a business and information processing management system needs to be created that not only identifies similar transactions to known fraud cases but also keeps up with how criminals change their habits by recognizing new fraud cases.

The ACFE and the Association of Certified Money Laundering Specialists have documented fraud typologies \citep{ACAMS2019}. These roadmaps are useful for understanding the crime and electronic fraud patterns. Further, they are useful for ascertaining which data fields must be added to the data set, in addition to the fields that must be developed to optimize the predictive capabilities of the machine learning system to increase accuracy while reducing false positives and negatives.

\subsection{Research Objective}

This is an investigative study designed to build a baseline foundation for future work in the application of machine learning for fraud detection.  It is not a survey of all approaches, although the groundwork is supported by a vast survey of available research in machine learning for classification and resampling methods in handling the principle issue facing the study of fraud: how to deal with imbalanced data sets? The contributions of this study help answer the following research questions: 

\begin{enumerate}
    \item Do stacked generalizations in combination with resampling methods improve the identification of rare events in fraud data sets? 
    \item If so, which stack generalizations’ resampling combinations work better than others?
\end{enumerate}

In typical machine learning efforts, the stakeholders define the problem and provide data to the data science team. The modeling output is then returned to the stakeholders for analysis. In the case of fraud, the model that best generalizes the data set may not be finding meaningful new cases of fraud. This work explores how information processing and management solutions ensure the predictive modeling results reflect the correct classification matrix values. 

This paper is organized as follows. In Section \ref{sec:Lit}, we review the relevant literature related to handling imbalanced data sets, to include discussion around the application of stacked generalizers and the contributions of this study. Section \ref{sec:Meth} describes the methodology used in the study, particularly around methods for handling imbalanced data sets at both the data-level, algorithm-level, cost-sensitive learning, and ensemble methods. We also discuss the evaluation metrics employed along with the computational experimentation. In Section \ref{sec:Res}, we provide the results of our stacked generalizations using resampling methods for fraud detection, and in Section \ref{sec:DiscNext} we discuss our findings as well as next steps. We conclude our work in Section \ref{sec:Conc} and pose some future research directions.

\section{Related Work}\label{sec:Lit}

Handling imbalance is not a matter of simply separating classes.  Imbalance requires consideration of the data structure, small data set size with small minority samples that cause overfitting and deficient generalization \citep{Yusof2017}, data quality, small disjuncts as a result of sparsity of data \citep{Japkowicz2001}, weak separation and class overlap at the boundary \citep{Kotsiantis2006}, feature space sparseness caused by the imbalance of classes \citep{Chawla2005}, `divide and conquer' class separation may cause data fragmentation \citep{Friedman1992}, detection of rare events may have a negative influence causing noise \citep{Maalouf2015}, and imbalance predisposes the minority class to small disjuncts that are more disposed to poor performance and errors caused by bias \citep{Maalouf2015}, noise, missing attributes and data size \citep{Maalouf2015}.  Another opinion regarding class imbalance is that `volume, complexity of data, class imbalance, concept drift, class overlap, and class mislabeling' are challenges.  `Data quality and detection accuracy' are closely linked, requiring the management of feature selection while keeping the detection rate high and reducing false positives \citep{Chen2018}.  

If the initial distribution method is not optimal, then which is? Weiss and Provost \citep{Weiss2003} addressed this issue in a thorough but not exhaustive study when determining which class distribution to select for the reduced training set, which was a condensed version for data access and cost sensitive purposes. Their testing showed that the minority class had a disproportionate number of errors and determined that a sampling strategy with the initial data set measured by accuracy and a re-balanced version using the receiver operating characteristic (ROC) curve performing as well but not worse.  Chawla \citep{Chawla2003} agreed with their assessment that the best distribution may not be the original and noted example sampling strategies \citep{Chawla2005} including random over- and under-sampling  methods, Tomek Links, Condensed Nearest Neighbor (CNN), Neighborhood Cleaning Rule (NCL), SMOTE and its variants SMOTETomek and SMOTEENN, cost-sensitive learning (CSL) and Ensembles using AdaBoost and SMOTEBoost. The essential problems to solve were resampling strategies and between versus within class imbalance \citep{Chawla2005}. 

The treatments for managing imbalanced data sets include handling the rebalancing at the data and algorithm levels, using CSL, feature extraction, ensemble routines and a hybrid combination of methods.  One of the first data-level resampling techniques used random sampling to either undersample the majority class (losing valuable data), oversample the minority class (duplicating observations) or a combination of both \citep{Chawla2005}. Data-level resampling rebalances the ratio between minority and majority classes since their disparity in numbers has a higher misclassification cost for minority samples than the majority, which leads to metric evaluation errors \citep{Chawla2002}. Classification of the minority class does not improve with the duplication of samples in random oversampling \citep{Ling1998,Japkowicz2000,Chawla2002}.  A discovery using a more intelligent and focused approach called Synthetic Minority Over-sampling TEchnique (SMOTE) skews the minority oversamples just slightly, making the decision boundary between classes less specific and, thus, improving predictive performance \citep{Chawla2002}. This technique can be used to undersample the majority with a similar affect. 

A means to balance the cost of misclassifying the minority class without the use of resampling is by CSL, which makes the cost of misclassifying the minority class greater than the majority rebalancing the relationship \citep{McCarthy2004}.  This study found that random over- and undersampling performed similarly to CSL. A study by Maalouf also found that CSL and resampling achieved similar results \citep{Maalouf2015}.  Weiss found that CSL performed better with more data \citep{Weiss2007}.  Examples are MetaCost, AdaCost, CSBI, CSB2, RareBoost, AdaC1, AdaC2, and AdaC3. Improvements in feature selection can be made by association rules rather than correlation alone \citep{Rajeswari2015}.  This improves performance by reducing the data set.  Reducing dimensionality may also lose invaluable data that can be combined using association rules which reduces noise \citep{Chen2018}.  Zheng \citep{Zheng2004} found that imbalanced data sets are not suited to traditional feature selection and instead also proposed combining features. Ensemble applications includes more advanced cross-validation techniques that incorporate more successful model selection approaches \citep{Wolpert1992}. 

Ting and Witten \citep{Ting1999} noted that class probabilities are a fundamental requirement for applying stacked generalizers, which were found to be better performers than single classifiers.  Methods with data preprocessing include RUSBoost, SMOTEBoot, SMOTEBagging, EasyEnsemble, BalanceCascade, RareBoost and EUSBoost. Hybrid combinations appear to compensate for the weaknesses inherent in a single routine.  Many algorithms enable a cost weighting parameter to be set.  Examples are cost sensitive ensemble \citep{Zhang2013}, cost sensitive support SVM and QBC using SVM, C.45, NB, and Adaboost. There are often different and opposing outcomes documented in research conclusions which may be a reflection of the differences in the data set structure, source, size, variances in development decisions and the combination of data sets used when the differences are not significant. Sesmero et al. \citep{SesmeroM.Paz2015Geoh} also noted many contradictions in research results with no consensus regarding optimal stacking parameters.  They believed these could be reasons why the technique is used less often than straight forward bagging and boosting in real-world problems.  The complex evaluation process, as well as high computational costs, contribute to reasons why the technique is less often applied. Classification problems can be completed off-line, making the higher overhead for stacking still a viable option. 

Naimi and Balzer \citep{NaimiAshley2018Sgai} note that stacking is increasingly used in epidemiology, but it is observed that prediction is separate from the recognition of causal effects.  The complexity of algorithms and stacks of algorithms is opaque and considered a black box rather than an identification of causal relationships. The unknowns that are not distinguished by the meta learner in the data set include "confounders, instrumental variables, mediators, colliders, and the exposure."  Therefore, one needs to carefully think through the causal structure rather than taking the results at face value. Wong et al. \citep{WongManLeung2020Ceos} used a cost-sensitive neural network ensemble method of Stacked Denoising Autoencoders on six data sets, as well as Cost-Sensitive Deep Neural Network (CSDNN) and Cost-Sensitive Deep Neural Network Ensemble (CSDE) models against several typically used machine learning algorithms, including: Logistic Regression (LR), Neural Network (NN), Support Vector Machine (SVM), Bayesian Network (BN), Decision Tree (DT), boosting algorithms (LogitBoost, AdaBoost, RUSBoost, RBBoost, SMOTEBoost, Bagging, Voting, and BalanceCascade, and cost-sensitive algorithms (AdaCost and MetaCost). The test results showed that the CSDE and CSDNN models performed first and third best, respectively, for the True Positive Rate (TPR), True Negative Rate (TNR) and area under the curve (AUC) metrics in average rankings, as well as low generalization gaps between training and test results.

Increasingly, stacked generalization in conjunction with data-level resampling is used as the method of choice in scientific applications and where machine learning is being applied to improve generalization in the data set. Given the work of Chawla \citep{Chawla2002} that developed the SMOTE technique, this study develops a similar solution for fraud detection by leveraging the k-nearest neighbor (KNN) method \citep{Winston2009} and applying mathematical formulas that identify the inherent characteristics of fraud data using distance formulas. Further, Wang et al. \citep{wang2012extract} studied the automated feature selection processes, which found that the quality of the data set and its subsequent performance are better matched with the actual object of prediction. Fraud investigators can better find the best features based on the fraud incident, while confirming that the necessary data is contained in the data set. The automated feature selection algorithm assumes that all necessary data is present and the best-performing algorithm would determine the best features. This study assumes that all the necessary data is not present and must be confirmed through reverse engineering the actual fraud event. An automated feature selection would improve prediction after the confirmation and validation process by fraud investigators. Finally, Ting and Witten \citep{Ting1997} conclude that multiple classifiers are better than single classifiers, which is similar the findings in this study.

\section{Methodology}\label{sec:Meth}

The scarcity of publicly available fraud data sets is one of the most significant hurdles to overcome in the development of machine learning models for fraud detection. This study investigates stacked generalizations on fraud imbalanced data sets with resampled data. Given the unique characteristics of fraud data relative to other publicly available two-class data sets, including other than fraud data would not serve the research conditions.  Although fraud typology patterns may repeat, fraud data may not be systematic nor orderly and can follow complex deceptive schemes to commit the crime, as well as obfuscate the identity of the perpetrator.  Publicly available two-class data sets generally are sourced from science studies and have complicated systematic and/or orderly patterns. The binary classification separation of the distributions into classes would be relevant, but the relationships in the data structure and metrics from the resampling and predictions would be less applicable.

Rather than focus on experiments across data sets, the study strategy is centered on a within the data set comparative analysis tactical approach.  A single data set is used to delve deeper into changes and relationships that arise in preprocessing through the algorithm prediction phases.  Only one data source met the study requirements, which was a credit card fraud data set (creditcard.csv) from the Machine Learning Group at the Université Libre de Bruxelles in Brussels, Belgium \citep{MLG2004}. The data set has 284,807 entries of which 492 are known fraudulent and 184,315 are normal; this has an extreme imbalance ratio (IR) of 0.17\%, making its development results directly applicable to the main investigative study objective. Because only a single data set was examined in this study, no conclusions can be drawn until more fraud data sets are studied in future research efforts.

\subsection{Handling Imbalanced Data Sets}

In this study, a combination of data-level resamples (in addition to a full data set) and 11 algorithm classifier pairs were used to handle class imbalance for a combination of 88 baseline models. The algorithm descriptions are organized first under their more specific characteristics, such as CSL and ensembles, with the residual examples documented under algorithm applications. 

\subsubsection{Data-Level}

Data-level imbalance handling happens at an initial phase to rebalance the class imbalance ratios, which then go through the algorithm level handling phase. Seven resampling techniques were chosen for prepossessing; random over- and undersampling , SMOTE, SMOTEENN, BorderlineSMOTE, SMOTETomek and ADASYN, in addition to using the full data set for comparison purposes. Random oversampling keeps the majority class and randomly oversamples the minority class by the class difference with replacement.  The result is that the new minority samples are duplicates of existing samples causing overfitting. Random undersampling keeps the minority class and randomly undersamples (removing entries) from the majority class losing valuable data in the routine, which can be done with or without replacement.  

The SMOTE technique oversamples the minority using a more intelligent sampling application that improves prediction using KNN classification \citep{Chawla2002}.  It is considered a Lazy method since it does not model the training data but uses both training and test data in the `just in time' prediction.  In KNN, an odd number for $K$ is selected in order that the voting algorithm always has a winner. The first point is chosen from which the $K$ number of neighbors is calculated of each minority class feature vector by the shortest distance.  Then, each point votes positioned on its own class with the majority vote, deciding which class the initial point belongs. This information is kept in an array for the next step.

The minority class is then oversampled, with the new synthesized minority vector not being a duplication of previous entries and, thus, reducing overfitting. The number of oversamples, with the majority class less than the minority class, are calculated, and each new sample is taken from the distance of the line segment between the minority class feature vector sample selected and its nearest neighbor (from the stored array) measurement. This new sample is then multiplied by a random number between 0 and 1. These new synthetic samples create a less specific and general decision boundary line improving classification prediction. The higher the $K$ value the more even the decision boundary. The lower the $K$ the rougher the decision boundary. 

There are many different distance measures that can be applied\footnote{Distance measurements include Euclidean, Mahalanobis, Haversine, Hamming, Canberra, BrayCurtis, Jaccard, Matching, Dice, Kulsinski, RogersTanimoto, RussellRao, SokalMichener, and SokalSneath, Hellinger Distance.} with the most common being the Euclidean Distance, which is the ordinary straight-line distance between two points $(x\textsubscript{1},y\textsubscript{1})$ and $(x\textsubscript{2},y\textsubscript{2})$. Other possible options are building KNN from scratch by means of a similar SMOTE-like approach using another distance measurement.  The optimal distance measure for a data set may be determined by the data structure and mathematical relationships of the data at a more abstract plane.
 
For example, in a National Health Institute study \citep{Hu2016} the Euclidean, cosine, Chi square, and Minkowsky distance measures were compared and found that the Chi square distance performed better for medical domain data sets. Sentimentality prediction found that cosine improved the discovery of boundaries between classes \citep{Winston2010}. The optimal distance measurement for fraud transactions has not been explored and is very challenged by the lack of systematic and organized ways fraud is committed and the deficiency of publicly available data sets for researchers to access for their work.  On reflection, finding that cosine supports sentimentality prediction mathematical relationships would not have been an obvious choice because of the complexity and boundless possible ways of writing text.  Therefore, the possibility of recognizing a core mathematical relationship for fraud typologies is quite promising.
 
The other intelligent resampling techniques implemented in this study, SMOTETomek, SMOTEENN, BORDERLINESMOTE, and ADASYN, are explained in greater detail in Table \ref{tab:resampling}.

\begin{center}
\begin{table}[pos=hbt!]
\begin{tabular}{ |p{5cm}|p{10cm}| }
\hline 
\FloatBarrier
\textbf{Resampling Technique} & \textbf{Description} \\
\hline
SMOTETomek & Combines SMOTE and Tomek Links: removes noisy and unnecessary points that are internal and farther from the boundary, preserving only samples close to the boundary line \citep{Tomek1976}. \\ 
\hline
SMOTEENN (Smote Edited Nearest Neighbor) & An under-sampling cleaning technique that removes anomalous noisy majority predictions that contradict the majority class along the decision boundary. \\
\hline
BorderlineSMOTE  & Over-samples the minority at the decision boundary, improving classification rather than sampling further from the decision boundary that contribute less to the classification prediction \citep{Han2005}.\\
\hline
ADASYN (Adaptive Synthetic Sampling Technique)  & Over-samples centered on the distribution of the minority samples, reducing bias and modifies the decision boundary toward those that are more difficult to learn \citep{He2008}.\\
\hline
\end{tabular}
\caption{\label{tab:resampling} Other Intelligent Resampling Techniques}
\end{table}
\end{center}

\subsubsection{Algorithm-Level}

Any number of algorithms can be applied with or without resampling techniques.  The test strategy in this study was to compare a diverse but not a comprehensive list of machine learning classifiers, including those based on probability (Naive Bayes), classification (Decision Tree - C4.5, KNN and SVM) and Neural Networks (MLP). Naive Bayes (GaussianNB) has its foundation in the Bayes rule named after Thomas Bayes (1702-1761), which provides the logic for conditional probability. The Naive Bayes algorithm can be built manually and is the ratio of two frequencies. 

Decision tree classifier examples include ID3, C4.5, and CART, which are the three fundamental decision trees. C4.5 was developed from ID3 (iterative Dichotomer 3); both were discovered by Ross Quinlan \citep{Quinlan1987} and build decision trees piece by piece in a top-down fashion, determining the best advantage \citep{Sharma2013}. KNN separates classes by first calculating the number of nearest neighbors then measures the distance from each point to its nearest neighbors to vote upon which class the point should belong. For a more detailed explanation, see SMOTE above.

Support Vector Machines seek to find a plane that separates the classes, which is a concept that was first introduced by Vladimir Vapnik Ph.D in 1990 \citep{James2013}. Twenty years after it was introduced, it is commonly accepted as one of the best classification methods.  The goal is to design a hyperplane to separate a binary classifier labeled as positive or negative points. The MLP Classifier is a multi-layer perceptron classifier neural network back propagation approach that learns the structure in the data without first imposing assumptions on the data.  The learning by the hidden belief layer is enhanced in multiple layers in a progressive recursive function to gain learning patterns successively from the data \citep{Ng2016b}.

\subsubsection{Cost-Sensitive learning}

Handling imbalanced data sets is essentially a weighting problem that has been solved with straightforward, as well as creative, resampling solutions.  Cost-sensitive learning uses weighting on the full imbalanced data set, which has been found to be very effective. The cost of misclassification on the minority class is weighted or penalized more heavily than a correct classification.  The true costs include the costs of misclassification, testing and human intervention \citep{Turney2000,Shilbayeh2015}, as represented in the Cost Matrix (Table \ref{tab:cost}) initially developed by Hand \citep{Hand2007,Bahnsen2018}. The administrative costs of a false positive (C\textsubscript{FP\textsubscript{i}} = C\textsubscript{admin}) are equal to the true positive (C\textsubscript{TP\textsubscript{i}} = C\textsubscript{admin}), since in the latter instance the account holder needs to be contacted.  The cost of a false negative (C\textsubscript{FN\textsubscript{i}} = 100C\textsubscript{admin}), missing the fraud, is significantly greater and represented by the actual loss of the transaction amount (C\textsubscript{FN\textsubscript{i}} = A\textsubscript{MT\textsubscript{i}}). 

\begin{center}
\setlength\tabcolsep{0pt}
\begin{table}[pos=hbt!]
\centering 
\begin{tabular}{c >{\bfseries}r @{\hspace{0.7em}}c @{\hspace{0.4em}}c @{\hspace{0.7em}}l}
  \multirow{10}{*}{\rotatebox{90}{\parbox{1.1cm}{\bfseries\centering \vspace{.5cm}Predicted\\ \vspace{.5cm}}}} & 
    & \multicolumn{2}{c}{\bfseries Actual} & \\
  & & \bfseries y\textsubscript{i}$=1$ & \bfseries y\textsubscript{i}$=0$ & \bfseries  \\
  & c\textsubscript{i}$=1$ & \MyBox{CTP\textsubscript{i}}{=C\textsubscript{admin}} & \MyBox{CFP\textsubscript{i}}{=C\textsubscript{admin}}  \\[2.4em]
  & c\textsubscript{i}$=0$ & \MyBox{CFN\textsubscript{i}}{=AMT\textsubscript{i}} & \MyBox{CTN\textsubscript{i}}{=0} \\
\end{tabular}
\caption{\label{tab:cost}Cost Matrix: Cost of Misclassification}
\end{table}
\end{center}

There are two techniques of CSL that can be applied through machine learning.  The direct method alters accuracy-based algorithms in the way that includes cost weighting for misclassification \citep{Lomax2013a,Shilbayeh2015}. Since the algorithm is developed in a way that rewards lower costs, raising the weight of misclassification will over correct and, therefore, correctly classify. Weights of misclassification that are higher for the false positives than false negatives vary inversely as the class size increases. The indirect method applies a wrapper to an algorithm effectively altering how the misclassification is weighted \citep{Shilbayeh2015}. One primary issue is that the initial weighting cost is unknown requiring that a starting point be determined.  

For example, a direct method cost sensitive tree decides how to build branches being controlled by entropy introduced by an information cost factor in selecting how to add attributes \citep{Shilbayeh2015}. Information theory, usually referenced as entropy, quantifies irregularity and measures class distribution consistency \citep{Shannon1948,Quinlan1987,Lomax2013b}. In this study, the Random Forest algorithm was used with a cost-sensitive weighting parameter. CSL can be used on the whole data set, as well as with resampling, which can be an evaluation point to check whether preprocessing improves performance.

\subsubsection{Ensemble Methods}

Ensemble methods can be used with classifiers (\textit{AdaBoost}, \textit{BaggingClassifier}, \textit{EasyEnsemble}, \textit{RUSBoost}, \textit{GradientBoostingMachine}, etc.), with tacked Generalizers, or with other hybrid approaches.  They are often referred to as a mixture of experts that include cross-validation and voting strategies to improve prediction outcomes. The \textit{BaggingClassifier} is an ensemble meta-estimator that divides the sample into randomly selected groups with replacement.  Selection strategies include cross-validation with Bagging that uses majority voting rather than a winner-take-all routines. The predictions are combined \citep{Efron1979,Wolpert1992}, which may overfit with large trees \citep{Kraus2014}. In \textit{Boosting}, weak classifiers with higher error rates are combined with reweighting with voting that makes it a strong classifier \citep{Ray2015}. Strong and weak classifiers are measured by their error rates, with the former with a rate close to zero and a 100\% prediction accuracy and the latter an error rate less than or equal to the chance of a 50:50 outcome \citep{Winston2009}. 

The goal is to find a solution that reduces variance and, therefore, does not overfit and does not increase bias which can not resolve difficult machine learning challenges \citep{Winston2010}. Weak and strong classifiers may not always work in the same consistent way in the same areas.  By first categorizing the strong classifier with the lowest error rates than a weak classifier, boosting can find weights to improve performance through maximizing the likelihood of the inverse of the weights equalling one \citep{Winston2009} resulting in the reduction of overfitting \citep{Winston2010}.

Adaptive Boosting (\textit{AdaBoost}) works in a similar way in that CSL adjusts weights of misclassification with correct classification and, in fact, the boosting mechanism was developed to adapt CSL with boosting strategies \citep{Yin2013} incorporating weighted majority vote deciding the result. It can be used on the whole data set, which may be an evaluation point to check whether using \textit{AdaBoost} improves performance. \textit{GradientBoostingMachine} (GBM) is a boosted neural network classifier that makes no assumptions about the data (i.e., it is non-parametric) and bases the outcome from relationships learned in the modeling process. GBM builds on the boosting techniques using gradient descent to find the local minimum of the function and then builds new base learners iteratively to improve accuracy \citep{Natekin2013}. 

\textit{EasyEnsemble} uses random undersampling of the majority class that iterates through the samples first training then combining the model predictions \citep{Liu2009}. \textit{RUSBoost} applies random undersampling of the majority class before using boosting techniques of weighting then majority voting. \textit{RUSBoosting} is a simpler sampling technique compared to \textit{SMOTEBoost}, which applies the more complex SMOTE technique with the KNN classifier \citep{Seiffert2010}. \textit{EasyEnsemble} and \textit{RUSBoost} both use bagging as the principal approach and \textit{AdaBoost} as the classifier \citep{Liu2009}.

The Stacked Generalizer is a hybrid method introduced by David Wolpert to minimize the error rate using ensemble stacking to learn the bias in the initial individual classifiers with the output of the initial models \citep{Ting1997}. This is referred to as level-0 which then provide the input into the level-1 meta-learner phase that corrects for the bias identified using cross-validated prediction improving majority voting or weighting rather than using the winner-take-all strategy, thereby, enhancing the accuracy of the classifiers being stacked \citep{Wolpert1992}.

\begin{figure}
  \centering
  \includegraphics[scale=.45]{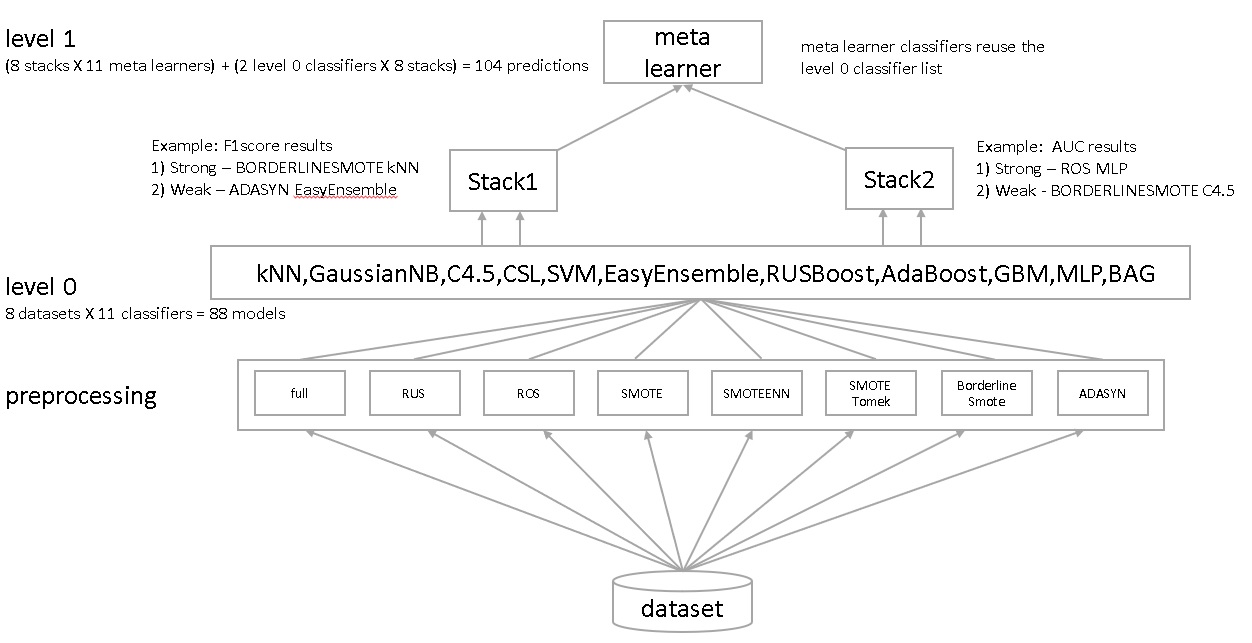}
  \caption{\label{fig:stacked}Stacked Generalization Machine Learning Architecture}
\end{figure}

Stacking ensembles combine bagging and boosting with different algorithms, in which the strengths and weaknesses in each model are weighted, thus promoting the weak classifiers and demoting the stronger classifiers in each resampling round. This focuses the algorithm on fixing misclassifications with only the improvements in accuracy included \citep{Ray2015}.  Rather than using a single result from an algorithm, stacked generalizations (see the architecture depicted in Figure \ref{fig:stacked}) combine methods results improving on the bias of each in the collection \citep{Wolpert1992}.  The bias of the individual classifiers are learned and then filtered out in the subsequent level-1 phase \citep{Wolpert1992}.  Wolpert noted that the rules for selecting level-0 and level-1 classifiers was unknown at the time of his writing in 1992. Yan and Han \citep{Yan2018} noted that stacking generalizations are not usually applied to imbalanced data sets, which is the core subject matter of this study. 

\subsection{Evaluation Metrics}

A machine learning model learns the patterns from the training set, which then makes inferences on test data to determine whether the new instances match these patterns (see the classification via the confusion matrix in Table \ref{tab:conf}).  The actual and predicted outcomes are categorized as true positive (both occurred and predicted), false positive (did not occur but was predicted, which is an incorrect prediction), a true negative (neither occurred nor predicted) and false negative (occurred but not predicted, representing an actual fraud that was missed by the machine learning classifier).  The numbers of each category are aggregated to build meaningful statistics.  The classification matrix also provides a threshold variable parameter greater than zero to 100\% with the default being 50\% or the prediction being no better than chance.  The higher the threshold the fewer predictions will be included in the matrix though with more applicable predictions.

The most commonly used metric is the scalar accuracy value, (TP+TN)/(TP+TN+FP+FN), which is not useful for imbalanced data sets since the minority class is under represented and is not a good measure of classifier performance \citep{Fawcett1997,Fawcett2005} since the misclassification costs of false positives (cost of a fraud investigator reviewing the event) and negatives (total loss of the transaction if the FN is found to be accurate) are not included.  The AUC, an array of the entire threshold range of accuracy, is more effective and efficient in gauging classifier performance. Once the data is rebalanced, accuracy becomes more useful.

\begin{center}
\begin{table}[pos=hbt!]
\centering
\begin{tabular}{c >{\bfseries}r @{\hspace{0.7em}}c @{\hspace{0.4em}}c @{\hspace{0.7em}}l}
  \multirow{10}{*}{\rotatebox{90}{\parbox{1.1cm}{\bfseries\centering \vspace{.5cm}Predicted\\ \vspace{.5cm}}}} & 
    & \multicolumn{2}{c}{\bfseries Actual} & \\
  & & \bfseries y\textsubscript{i}$=1$ & \bfseries y\textsubscript{i}$=0$ & \bfseries  \\
  & y$'=1$ & \MyBox{True}{Positive} & \MyBox{False}{Positive}  \\[2.4em]
  & y$'=0$ & \MyBox{False}{Negative} & \MyBox{True}{Negative}
\end{tabular}
\caption{\label{tab:conf}Machine Learning Model Classification Matrix}
\end{table}
\end{center}

Precision and recall, which are used to build the F1-score, are more applicable to the unequal class distributions in skewed data since they include false positives and negatives. Precision measures the percentage of correct predictions with respect to the total correctly predicted, determining when correctly predicted and how often and measuring the exactness. High precision is preferred, and it should be used when the cost of FPs are high, such as through reputation loss of falsely accusing a customer of fraud. Precision is used as the y-axis in the Precision-Recall curve. Recall (also known as sensitivity and TP Rate, or TPR) calculates the proportion of positives correctly predicted; a low recall is preferred.  Recall is used as the x-axis in the Precision-Recall curve and y-axis in the ROC curve. The difference between precision and recall are FP and FN in the numerator. FP have associated investigative costs that can be interpreted to be direct overhead costs, and FN are opportunity costs of having missed the prediction which is a loss of the value of the business event that could have been spent elsewhere. Quantifying the direct and opportunity costs provides a more consequential means of discovering the weight of the precision versus recall trade-off \citep{Accumulation2018}.
 
F1-score calculates the trade-off between precision and recall that is moderated by using a threshold greater than 0 to 100\%. The higher the precision (if correctly predicted, how often and exact) the lower the recall (correct predictions). The preference is a F1-score of 1 with a high precision and high recall, which identifies the optimal balance including with imbalanced data sets \citep{Ng2016a}. FP Rate (FPR) computes incorrect predictions or the positive predictions incorrectly classified. FPR is used in the x-axis of the ROC curve. The AUC quantifies the ROC performance and summarizes accuracy for the entire test result outcomes, which represent the threshold range rather than using just the scalar accuracy value \citep{Fawcett2005}.  The visualization is a graph with TPR on the y-axis and FPR (i.e., 1 - specificity) on the x-axis. The optimal cutoff is where sensitivity is relatively equal to Specificity. The diagonal line from (0,0) to (0,1) is the 50 percent threshold representing random selection. AUC is indifferent to skewed data and can be summarized using a scalar AUC value \citep{Fawcett2005}.
 
Although some metrics may not be relevant or useful for evaluating classifier performance on imbalanced data sets, the prehandling technique of resampling balances the data distribution and modifies the original shortcomings of these measurements, though perhaps not mitigating them fully since the change in balance is not adding new data as much using creative techniques to equalize the number of observations in each class. Before the data is rebalanced, the AUC may be overly optimistic \citep{He2009}; therefore, the Precision-Recall curve is a better tool for evaluation of machine learning classifier performance on skewed data providing more discriminating information.

\subsection{Computational Experimentation}

A reuseable and reconfigurable test harness was built with viable algorithm and sampling options to evaluate a variety of resampling plans, baseline machine learning classifiers, and meta learners (each of which can be easily substituted) help keep a broad range of combination options available to investigate the best production configuration for the top project strategy. Testing consists of: 1) preprocessing for resampling of the data set; 2) modeling and predicting the level-0 baseline classifier resample pairs; and 3) building stacks from the level-0 findings to be predicted with the level-1 meta learners. Eight data samples were teamed with each of 11 machine learning classifiers totaling 88 level-0 baseline models, using the following resampling strategies and classifiers (see Table \ref{tab:stacks}). After the baseline level-0 models were predicted, their outcomes were sorted by AUC and F1-score to find the strongest and weakest classifier resample pairs with 8 stacks developed for test purposes.

\begin{itemize}
\item Resampling Mechanism: full, ROS, RUS, SMOTE, SMOTETomek, SMOTEENN, BorderlineSMOTE, ADASYN
\item Machine Learning Classifier: KNN, GaussianNB, C4.5, CSL, SVM, EasyEnsemble, RUSBoost, AdaBoost, GBM, Bagging and MLP
\end{itemize}

\begin{center}
\begin{table}[pos=hbt!]
\begin{tabular}{ |p{1.2cm}|p{.75cm}|p{5.5cm}|p{5.5cm}| }
\hline 
\textbf{Metric} & \textbf{Stack} & \textbf{Classifier 1} & \textbf{Classifier 2}\\
 \hline
 F1-score & \#1 & w-ADASYN EasyEnsemble & s-BORDERLINESMOTE kNN \\ 
 F1-score  & \#2 & w-BORDERLINESMOTE RUSBoost & w-SMOTE GaussianNB\\
 F1-score  & \#3 & s-SMOTEENN MLP &s-SMOTE CSL \\
 F1-score  & \#4 & tree-ROS CSL & tree-RUS C4.5\\
 AUC & \#5 & w-BORDERLINESMOTE C4.5 & s-ROS MLP \\ 
 AUC  & \#6 & w-ADASYN BAG & w-SMOTE SVM\\
 AUC  & \#7 & s-SMOTETomek GaussianNB & s-SMOTEENN AdaBoost \\
 AUC  & \#8 & tree-ROS C4.5 & tree-full CSL\\
 \hline
\end{tabular}
\caption{\label{tab:stacks}Stacks of Classifier and Resample Pairs from Level-0 Models}
\end{table}
\end{center}
\FloatBarrier

These eight stacks were remodeled against each of the 11 meta learner classifiers with the original baseline entries also present in the level-1 phase for contrast, which totaled (88 + 16) 104 predictions. The results of the level-0 and level-1 predictions were analyzed and then combined to see if and where improvements were obtained with data resamples and classifiers, and whether level-1 predictions provided notable benefits reducing false positives and negatives while increasing true positives.  The final outcome was sorted in order by AUC, F1-score and accuracy.

\section{Modeling Results}\label{sec:Res}

The model development findings were reviewed in three phases: 1) level-0 baseline, 2) level-1 stack generalizations with meta learners, and 3) combining both level-0 and level-1. From Table \ref{tab:level0}, the top 17 entries produced an AUC of 96\%, of which 13 utilized more informed resampling techniques (ADASYN, SMOTE and SMOTE variants), one full data set, and three ROS/RUS resamples.  The top six entries had very similar metrics. The top three predictions were applied with the MLP classifier with a SMOTE/ BORDERLINESMOTE or random oversampling (ROS) approach, each returning very similar metrics.  The ROS MLP entry would most likely be chosen for production on cost factor alone, since the random oversampling technique has a lower computation processing and time overhead.

\begin{center}										
\begin{table}[pos=hbt!]									\begin{tabular}{|p{3cm}|p{2.5cm}|p{.5cm}|p{.7cm}|p{.4cm}|p{1cm}|p{1.1cm}|p{0.6cm}|p{1.2cm}|p{.6cm}| }				
\hline 
\textbf{Test Run} 	&	\textbf{Classifier}	&	\textbf{TP}	&	\textbf{FP}	&	\textbf{FN}	&	\textbf{TN}	&	 \textbf{Accuracy}		&	\textbf{F1 Rank}	&	 \textbf{F1-score}	&	\textbf{AUC}	\\				 
\hline			
0SMOTE	&	MLP	&	142	&	20	&	43	&	113718	&	0.9994		&	8	&	0.8184	&	0.96	\\
\hline 
0ROS	&	MLP	&	142	&	28	&	41	&	113712	&	0.9994		&	9	&	0.8046	&	0.96	\\
\hline 
0BLSMOTE	&	MLP	&	150	&	27	&	46	&	113700	&	0.9994		&	10	&	0.8043	&	0.96	\\
\hline 
0BLSMOTE	&	GBM	&	147	&	28	&	49	&	113699	&	0.9993		&	11	&	0.7925	&	0.96	\\
\hline 
0SMOTEENN	&	AdaBoost	&	131	&	28	&	65	&	113699	&	0.9992		&	12	&	0.7380	&	0.96	\\
\hline 
0BLSMOTE	&	AdaBoost	&	127	&	28	&	69	&	113699	&	0.9991		&	13	&	0.7237	&	0.96	\\
\hline 
0BLSMOTE	&	GaussianNB	&	171	&	671	&	25	&	113056	&	0.9939		&	14	&	0.3295	&	0.96	\\
\hline 
0SMOTETomek	&	GaussianNB	&	171	&	694	&	30	&	113028	&	0.9936		&	15	&	0.3208	&	0.96	\\
\hline 
0RUS	&	GaussianNB	&	170	&	731	&	31	&	112991	&	0.9933		&	16	&	0.3086	&	0.96	\\
\hline 
0ROS	&	GaussianNB	&	160	&	697	&	23	&	113043	&	0.9937		&	17	&	0.3077	&	0.96	\\
\hline 
0full	&	GaussianNB	&	163	&	706	&	29	&	113025	&	0.9935		&	18	&	0.3073	&	0.96	\\
\hline 
0SMOTEENN	&	GaussianNB	&	169	&	735	&	27	&	112992	&	0.9933		&	19	&	0.3072	&	0.96	\\
\hline 
0ADASYN	&	GaussianNB	&	149	&	668	&	26	&	113080	&	0.9939		&	20	&	0.3004	&	0.96	\\
\hline 
0SMOTE	&	GaussianNB	&	155	&	695	&	30	&	113043	&	0.9936		&	21	&	0.2996	&	0.96	\\
\hline 
0BLSMOTE	&	EasyEnsemble	&	178	&	4874	&	18	&	108853	&	0.9571		&	22	&	0.0678	&	0.96	\\
\hline 
0SMOTEENN	&	EasyEnsemble	&	177	&	6876	&	19	&	106851	&	0.9395		&	23	&	0.0488	&	0.96	\\
\hline 
0ADASYN	&	EasyEnsemble	&	159	&	7304	&	16	&	106444	&	0.9357		&	24	&	0.0416	&	0.96	\\
\hline											
\end{tabular}	
\caption{\label{tab:level0}Level-0 Baseline With Highest AUC (Above 96\%)}
\end{table}								
\end{center}

The 104 predictions for the meta learners and stack elements displayed three predictions with AUC metrics of 97\% (see Table \ref{tab:level1}), which was followed by a group of 24 entries with an AUC of 96\%. Twenty-four of the top 27 AUC entries were meta learners with the best groups representing seven of the eight different stack combinations and the top AUC models representing three stacks types.

\begin{center}
\begin{table}[pos=hbt!]
\begin{tabular}{|p{1cm}|p{1cm}|p{1.2cm}|p{.7cm}|p{5cm}|p{5.5cm}|}
\hline
\textbf{Count} & \textbf{Type} & \textbf{Metric} & \textbf{Stack} & \textbf{Classifier 1} & \textbf{Classifier 2}\\ \hline
1 & meta & AUC  & \#6 & w-ADASYN BAG & w-SMOTE SVM\\
1 & meta & F1-score & \#1 & w-ADASYN EasyEnsemble & s-BORDERLINESMOTE kNN  \\
1 & meta & AUC  & \#7 & s-SMOTETomek GaussianNB & s-SMOTE\_ENN AdaBoost \\
\hline
\end{tabular}
\caption{\label{tab:level1}Top Level-1 by AUC (Above 97\%}
\end{table}
\end{center}

From the constructed combinations, the meta learners appear to have improved seven of the eight stacks (1, 2, 3, 5, 6, 7 and 8). The top entry had a reasonable FP statistic with the second and third having much higher FP values of 523 and 4918, respectively, creating a less likely scenario that these would be the selected model (see Table \ref{tab:level1meta}). The top three models had very different TP and FP values, suggesting uncertainty around which model should be selected that is representative of optimal performance while representing statistics that are meaningful for identifying fraud. 

\begin{center}
\begin{table}
\begin{tabular}{|p{3.5cm}|p{2.3cm}|p{.5cm}|p{.7cm}|p{.4cm}|p{1cm}|p{1.1cm}|p{1.3cm}|p{.6cm}| }					\hline \textbf{Test Run }	&	\textbf{Classifier}	&	\textbf{TP}	&	\textbf{FP}	&	\textbf{FN}	&	\textbf{TN}	&	 \textbf{Accuracy}	&	\textbf{F1-score}	&	\textbf{AUC}\\
\hline 
6metalearner	&	GBM	&	96	&	19	&	46	&	85281	&	0.9992	&	0.7471	&	0.97	\\\hline	
1metalearner	&	GaussianNB	&	122	&	523	&	22	&	84775	&	0.9936	&	0.3092	&	0.97	\\\hline	
7metalearner	&	EasyEnsemble	&	133	&	4918	&	13	&	80378	&	0.9423	&	0.0511	&	0.97	\\\hline	
5metalearner	&	GBM	&	116	&	17	&	40	&	85269	&	0.9993	&	0.8028	&	0.96	\\\hline	
7metalearner	&	MLP	&	108	&	16	&	38	&	85280	&	0.9994	&	0.8000	&	0.96	\\\hline	
1metalearner	&	MLP	&	108	&	18	&	36	&	85280	&	0.9994	&	0.8000	&	0.96	\\\hline	
7metalearner	&	GBM	&	109	&	19	&	37	&	85277	&	0.9993	&	0.7957	&	0.96	\\\hline	
5stackROS	&	MLP	&	140	&	27	&	48	&	113708	&	0.9993	&	0.7887	&	0.96	\\\hline	
8metalearner	&	MLP	&	106	&	21	&	40	&	85275	&	0.9993	&	0.7765	&	0.96	\\\hline	
1metalearner	&	GBM	&	105	&	22	&	39	&	85276	&	0.9993	&	0.7749	&	0.96	\\\hline	
2metalearner	&	MLP	&	102	&	19	&	42	&	85279	&	0.9993	&	0.7698	&	0.96	\\\hline	
8metalearner	&	AdaBoost	&	93	&	21	&	53	&	85275	&	0.9991	&	0.7154	&	0.96	\\
\hline	
6metalearner	&	MLP	&	90	&	22	&	52	&	85278	&	0.9991	&	0.7087	&	0.96	\\
\hline
2metalearner	&	AdaBoost	&	88	&	24	&	56	&	85274	&	0.9991	&	0.6875	&	0.96	\\
\hline	
2metalearner	&	GaussianNB	&	127	&	502	&	17	&	84796	&	0.9939	&	0.3286	&	0.96	\\\hline	
7metalearner	&	GaussianNB	&	124	&	485	&	22	&	84811	&	0.9941	&	0.3285	&	0.96	\\\hline	
5metalearner	&	GaussianNB	&	134	&	526	&	22	&	84760	&	0.9936	&	0.3284	&	0.96	\\\hline	
2stackSMOTE	&	GaussianNB	&	179	&	719	&	30	&	112995	&	0.9934	&	0.3234	&	0.96	\\\hline	
8metalearner	&	GaussianNB	&	127	&	521	&	19	&	84775	&	0.9937	&	0.3199	&	0.96	\\\hline	
6metalearner	&	GaussianNB	&	120	&	503	&	22	&	84797	&	0.9939	&	0.3137	&	0.96	\\\hline	
3metalearner	&	GaussianNB	&	125	&	547	&	20	&	84750	&	0.9934	&	0.3060	&	0.96	\\\hline	
7stackSMOTETomek	&	GaussianNB	&	158	&	727	&	29	&	113009	&	0.9934	&	0.2947	&	0.96	\\\hline	
3metalearner	&	RUSBoost	&	124	&	669	&	21	&	84628	&	0.9919	&	0.2644	&	0.96	\\\hline	
2metalearner	&	RUSBoost	&	127	&	1068	&	17	&	84230	&	0.9873	&	0.1897	&	0.96	\\\hline	
2metalearner	&	EasyEnsemble	&	131	&	4272	&	13	&	81026	&	0.9498	&	0.0577	&	0.96	\\\hline	
6metalearner	&	EasyEnsemble	&	130	&	5053	&	12	&	80247	&	0.9407	&	0.0489	&	0.96	\\\hline	
1metalearner	&	EasyEnsemble	&	128	&	5492	&	16	&	79806	&	0.9355	&	0.0445	&	0.96	\\	
\hline		
\end{tabular}
\caption{\label{tab:level1meta}Level-1 Meta Learner with Stack Predictions by AUC}
\end{table}
\end{center}

When combining the level-0 and level-1 predictions, the top three models were the same as the top level-1 predictions with an AUC of 97\%. In comparing the classification metrics of the models with 96\% AUC to the top 97\% AUC models, the top 11 96\% models had better FP statistics then two of the three 97\% AUC models. In addition, the 96\% AUC MLP ROS model used a single algorithm. As depicted in Table \ref{tab:level01}, the top 44 models had AUC values of 96\% or 97\% of which 24 were meta learners, 3 were stacks, and 17 were baseline single algorithms.

\begin{figure}
  \centering
  \includegraphics[scale=.8]{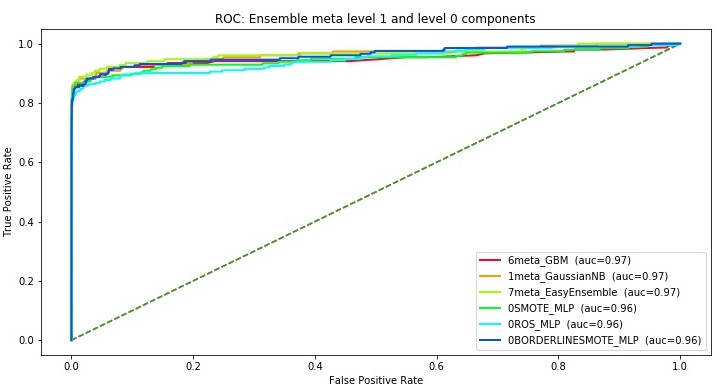}
  \caption{\label{fig:ROCcombined}ROC Curve of Combined Results}
\end{figure}

The meta learners with informed resampling methods continued to demonstrate optimal metric performance in a combined level-0 and level-1 comparative analysis, achieving the top three models AUCs of 97\% and improving on the best baseline single algorithm AUC of 96\% (see Figure \ref{fig:ROCcombined}). In this combined list, three models were baseline level-0 pairs in positions 4-6 that performed very closely to the top three meta learners by AUC with better FP statistics, which makes model selection convoluted.

A noteworthy observation comes from the top three performing models, which are all meta learners representing stacks 6, 1, and 7. Stack 6 was comprised of the two weakest classifier pairs from AUC values and the Ensemble GBM classifier. Stack 1 combined the strongest and weakest classifier pairs from the F1-score with GaussianNB. Stack 7 consisted of two strong classifier resample pairs by AUC with EasyEnsemble. Stacking by combining weaker classifiers based on AUC returns better performance, which proved true in this case with the second and third highest models, 1 and 7, having much higher FP statistics \citep{joshi2002predicting}.

The first and only time that the full data set was listed appears in position 44/104 for the level-1 phase and eleven times in the combined list starting at 31/192, which demonstrates that resampling did have a marked prediction improvement. Cost-sensitive learning did not perform well; however, this may be due to the classifier choice and/or parameters chosen since CSL does well in other stacking generalization studies.

\begin{center}
\begin{table}[pos=hbt!]
\begin{tabular}{|p{3.4cm}|p{2.3cm}|p{.5cm}|p{.7cm}|p{.4cm}|p{1cm}|p{1.1cm}|p{1.3cm}|p{.6cm}| }				
\hline \textbf{Test Run}	&	\textbf{Classifier}	&	\textbf{TP}	&	\textbf{FP}	&	\textbf{FN}	&	\textbf{TN}	&	 \textbf{Accuracy}	&	\textbf{F1-score}	&	\textbf{AUC}\\
\hline 
6metalearner	&	GBM	&	96	&	19	&	46	&	85281	&	0.9992	&	0.7471	&	0.97	\\\hline
1metalearner	&	GaussianNB	&	122	&	523	&	22	&	84775	&	0.9936	&	0.3092	&	0.97	\\\hline
7metalearner	&	EasyEnsemble	&	133	&	4918	&	13	&	80378	&	0.9423	&	0.0511	&	0.97	\\\hline
0SMOTE	&	MLP	&	142	&	20	&	43	&	113718	&	0.9994	&	0.8184	&	0.96	\\\hline
0ROS	&	MLP	&	142	&	28	&	41	&	113712	&	0.9994	&	0.8046	&	0.96	\\\hline
0BLSMOTE	&	MLP	&	150	&	27	&	46	&	113700	&	0.9994	&	0.8043	&	0.96	\\\hline
5metalearner	&	GBM	&	116	&	17	&	40	&	85269	&	0.9993	&	0.8028	&	0.96	\\\hline
7metalearner	&	MLP	&	108	&	16	&	38	&	85280	&	0.9994	&	0.8000	&	0.96	\\\hline
1metalearner	&	MLP	&	108	&	18	&	36	&	85280	&	0.9994	&	0.8000	&	0.96	\\\hline
7metalearner	&	GBM	&	109	&	19	&	37	&	85277	&	0.9993	&	0.7957	&	0.96	\\\hline
0BLSMOTE	&	GBM	&	147	&	28	&	49	&	113699	&	0.9993	&	0.7925	&	0.96	\\\hline
5stackROS	&	MLP	&	140	&	27	&	48	&	113708	&	0.9993	&	0.7887	&	0.96	\\\hline
8metalearner	&	MLP	&	106	&	21	&	40	&	85275	&	0.9993	&	0.7765	&	0.96	\\\hline
1metalearner	&	GBM	&	105	&	22	&	39	&	85276	&	0.9993	&	0.7749	&	0.96	\\\hline
2metalearner	&	MLP	&	102	&	19	&	42	&	85279	&	0.9993	&	0.7698	&	0.96	\\\hline
0SMOTEENN	&	AdaBoost	&	131	&	28	&	65	&	113699	&	0.9992	&	0.7380	&	0.96	\\\hline
0BLSMOTE	&	AdaBoost	&	127	&	28	&	69	&	113699	&	0.9991	&	0.7237	&	0.96	\\\hline
8metalearner	&	AdaBoost	&	93	&	21	&	53	&	85275	&	0.9991	&	0.7154	&	0.96	\\\hline
6metalearner	&	MLP	&	90	&	22	&	52	&	85278	&	0.9991	&	0.7087	&	0.96	\\\hline
2metalearner	&	AdaBoost	&	88	&	24	&	56	&	85274	&	0.9991	&	0.6875	&	0.96	\\\hline
0BLSMOTE	&	GaussianNB	&	171	&	671	&	25	&	113056	&	0.9939	&	0.3295	&	0.96	\\\hline
2metalearner	&	GaussianNB	&	127	&	502	&	17	&	84796	&	0.9939	&	0.3286	&	0.96	\\\hline
7metalearner	&	GaussianNB	&	124	&	485	&	22	&	84811	&	0.9941	&	0.3285	&	0.96	\\\hline
5metalearner	&	GaussianNB	&	134	&	526	&	22	&	84760	&	0.9936	&	0.3284	&	0.96	\\\hline
2stackSMOTE	&	GaussianNB	&	179	&	719	&	30	&	112995	&	0.9934	&	0.3234	&	0.96	\\\hline
0SMOTETomek	&	GaussianNB	&	171	&	694	&	30	&	113028	&	0.9936	&	0.3208	&	0.96	\\\hline
8metalearner	&	GaussianNB	&	127	&	521	&	19	&	84775	&	0.9937	&	0.3199	&	0.96	\\\hline
6metalearner	&	GaussianNB	&	120	&	503	&	22	&	84797	&	0.9939	&	0.3137	&	0.96	\\\hline
0RUS	&	GaussianNB	&	170	&	731	&	31	&	112991	&	0.9933	&	0.3086	&	0.96	\\\hline
0ROS	&	GaussianNB	&	160	&	697	&	23	&	113043	&	0.9937	&	0.3077	&	0.96	\\\hline
0full	&	GaussianNB	&	163	&	706	&	29	&	113025	&	0.9935	&	0.3073	&	0.96	\\\hline
0SMOTEENN	&	GaussianNB	&	169	&	735	&	27	&	112992	&	0.9933	&	0.3072	&	0.96	\\\hline
3metalearner	&	GaussianNB	&	125	&	547	&	20	&	84750	&	0.9934	&	0.3060	&	0.96	\\\hline
0ADASYN	&	GaussianNB	&	149	&	668	&	26	&	113080	&	0.9939	&	0.3004	&	0.96	\\\hline
0SMOTE	&	GaussianNB	&	155	&	695	&	30	&	113043	&	0.9936	&	0.2996	&	0.96	\\\hline
7stackSMOTETomek	&	GaussianNB	&	158	&	727	&	29	&	113009	&	0.9934	&	0.2947	&	0.96	\\\hline
3metalearner	&	RUSBoost	&	124	&	669	&	21	&	84628	&	0.9919	&	0.2644	&	0.96	\\\hline
2metalearner	&	RUSBoost	&	127	&	1068	&	17	&	84230	&	0.9873	&	0.1897	&	0.96	\\\hline
0BLSMOTE	&	EasyEnsemble	&	178	&	4874	&	18	&	108853	&	0.9571	&	0.0678	&	0.96	\\\hline
2metalearner	&	EasyEnsemble	&	131	&	4272	&	13	&	81026	&	0.9498	&	0.0577	&	0.96	\\\hline
6metalearner	&	EasyEnsemble	&	130	&	5053	&	12	&	80247	&	0.9407	&	0.0489	&	0.96	\\\hline
0SMOTEENN	&	EasyEnsemble	&	177	&	6876	&	19	&	106851	&	0.9395	&	0.0488	&	0.96	\\\hline
1metalearner	&	EasyEnsemble	&	128	&	5492	&	16	&	79806	&	0.9355	&	0.0445	&	0.96	\\\hline
0ADASYN	&	EasyEnsemble	&	159	&	7304	&	16	&	106444	&	0.9357	&	0.0416	&	0.96	\\\hline
\end{tabular}
\caption{\label{tab:level01}Combined Level-0 and Level-1 Predictions by AUC}
\end{table}
\end{center}									

\begin{figure}
  \centering
  \includegraphics[scale=.8]{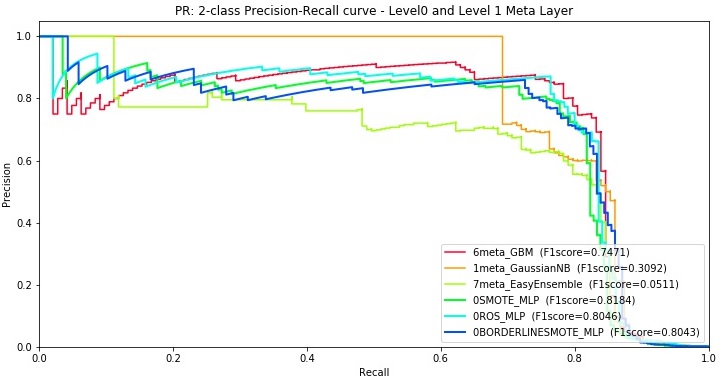}
  \caption{\label{fig:PRcurve}Precision-Recall Curve of Combined Results}
\end{figure}

Given the range of FPs with similar AUC scores, this demonstrates that the best machine learning model results are not definitive from statistics alone and need to be sent to fraud investigators for review; this will be the final determination of the success and failure of the machine learning modeling process.  This is primarily due to the unique conditions found in fraud data in that these crimes are committed through trickery and deceit, as well as being a moving target due to criminals constantly learning to game the system as soon as they play out each back door which closes when system controls are updated.  Choosing a model on statistics alone, without reviewing if the metrics actually match confirmed fraudulent transactions during at least an initial exploration and development effort with periodic reviews, would be a guessing based on numbers alone. As depicted in Figure \ref{fig:PRcurve}, the Precision-Recall curve demonstrates that the statistics within each test run show great variation within the individual classification statistics.

\section{Discussion}\label{sec:DiscNext}

The question is, which are the better predictions? After fraud investigation review, delving more deeply into the classification matrix metrics will provide a better understanding of the most useful model(s). Once the classification values are verified and the machine learning algorithms are rerun, leading to the selection of the best model, then the processing time and computational resources are evaluated.

The first step in the review of the model results incorporates the fraud investigator’s assessment of TP, FP and FN events for the top models, which is a necessary labor-intensive process before the final model selection. In reviewing the top performers of the combined final list, there is quite a difference in their classification statistics for the top AUC performers. The FP values range from 19 to 4918, which translates into a variance of 4899 more administrative case reviews for fraud investigators. The true and false positive results should be analyzed for why they failed to accurately predict, as well as all true positive findings should be manually evaluated and assessed for possible formal investigations being established to find whether a crime was committed and then identify who committed the crime. An interdisciplinary process comprised of the fraud investigation, accounting, database and data science teams can efficiently accomplish the continual feedback loop to support these investigations. As fraud investigators make improved decisions, changes may be made to the data set, variables, or observations to improve predictions so that the results are more meaningful rather than just having better generalization performance. 

A strong reciprocal relationship between the data science and fraud investigation teams creates a continual improvement update loop based on the initial briefing and debriefing from each team at the start and completion of their work in order to capture and update all lessons learned and best practices. Feedback from the investigation team may provide useful information for data scientists in how to improve the quality of the data set to support investigation outcomes. Machine learning is not the end goal but one excellent tool that is a link in a chain of many work efforts to help recognize fraudulent business operations.

When the best model is chosen a second step integrates cost-benefit analysis for computational performance. The smallest gains in statistics could mean serious savings to the next step in the process. Stacking generalizations have a development and processing overhead that need to be weighed between operation costs and predictive gains to justify their implementation in production. Each predictive gain may represent a future lower overhead in administrative costs associated with a fraud investigation review or an increase in fraud being successfully identified and/or prosecuted. Additional analysis of the new fraud cases found may include fraud event cost evaluation. Fraud events represented in a classification matrix should be viewed from a higher business perspective.  An incident of fraud is a statistic, but the value of each fraudulent transaction can vary within a substantial range of cost and/or damage.  

Even if the fraud investigators review does not justify using stacked generalizations for production, the cost and effort of building a test harness to evaluate a wide range of algorithm and sampling options to be used for research and development (R\&D) for project and process improvements may be justified.  This is when more formal studies on the effectiveness and efficiency of the machine learning process are conducted.  The data set is reviewed to see if the necessary attributes exist to find the fraudulent transactions, finding which resampling routines and classifiers are actually achieving results supported by the fraud investigator's conclusions, and when test strategies are explored to further improve machine learning performance.  The cost and effort of R\&D is weighed against yielding process improvements and updates to best practices that demonstrate reduced budgets for fraud investigator review and more successfully identified and/or prosecuted fraud cases.

\subsection{Next Steps}

There are several areas of focus with potential areas of study to improve predictive performance.  The first is a more theoretical inquiry into finding the best KNN distance measurement for sampling a fraud data set to optimize the classification process using informed resampling such as SMOTE or its variants. Another is to add more fraudulent events in the data set from known fraud cases in the business industry in an accumulative manner rather than using only existing current fraud transactions. 

This accumulative method can assist industry-wide solutions (using reverse engineering of the fraud event with case management, accounting, and other supportive fraud detection tools and team members) where the fraud typology can be outlined with possible system access points and routines including multiple event tracing.  This evidence can be placed in an overall fraud transaction framework then further divided by fraud type and industry for production.  Creating a knowledge base can reuse this information supporting businesses at an industry level rather than building solutions using only their own one-off experience.  A similar framework was recently successfully accomplished for managing data analytics in internal audit in partnership with the ACFE, Internal Audit Association (IIA), American Institute of Certified Public Auditors (AICPA) and private industry.  Internal audit had been overwhelmed by the volume and complexity of transactions being handled daily by businesses and agencies.  Only by creating a higher-level consortium to oversee and strategize at an industry wide perspective was a solution available.  This was a monumental effort and took approximately five years.  The exact same initial conditions are facing businesses and agencies responsible for identifying fraudulent events in their electronic systems.

The best practices and lessons learned from creating a framework and viable development solutions for internal audit can be leveraged to get an industry wide handle on fraud transactions.  Since internal and external audit have a due diligence requirement to build the finding and managing of fraudulent transactions into their work flow, there exists a strong possibility that the internal audit data analytics solutions have reusable solutions that already focus on fraud. These same internal audit data analytics solutions very likely are already in place in the same businesses and agencies that need improved identification of fraud solutions.  In some cases, the internal audit, fraud investigations, and machine learning data scientists are already working together.  Building a formal group and procedures for them to work together on an industry-wide strategy would be required before formal funding would be approved.  The low percentage of fraud being found along with the high percentage of revenue lost worldwide would justify the cost of investing in such a solution.

The internal audit data analytics initiative included financial transaction reviews. The problem space for identifying the requirements raised the work descriptions to a more abstract level which did not include confidential customer information. This means of solving a technical problem, without triggering prohibitions on information sharing across financial institutions, set an excellent precedent to guide the planning requirements for a similar project focused on building a framework around fraud typologies as they occur in business. A byproduct of having a consortium of businesses and professional practice organizations working together is the ability to piece together the transactions broken up by crime rings for the purpose of giving them a smokescreen for their criminal enterprise. An industry-wide framework would also be able to support cross-industry, federal and state agency, and law enforcement collaborations handling elaborate organized crime ring schemes that are committed between financial institutions and national borders. 

\section{Conclusion}\label{sec:Conc}

This study's focus was to discover the effect of stacked generalizations on resampled fraud imbalanced data and how to ensure the best machine learning model selection.  The work recognized an opportunity to discover a mathematical formula that would best generalize fraud data structure to be used in preferred resampling methods which is a theoretical solution.  Such a solution is hindered by the intrinsic issue that most fraud research is conducted within businesses not in a collaborative effort due to the confidentiality of the data; however, this perspective is changing and business consortiums are being established to work through industry solutions, although their data is not available to research groups where most of the theoretical breakthroughs in general are taking place.

Machine learning on fraudulent transactions has similar academic characteristics to the work being done in epidemiology. In neither case did automatically optimizing machine learning generalization necessarily find the rare events needed without further scrutiny.  Naimi and Balzer \citep{NaimiAshley2018Sgai} noted that the complexity of stacks obscured causal relationships in the input data and output outcomes.  As a result, a more thoroughly thought out understanding of their relationships needs to be the primary focus of the research rather than just a simple generalization performance improvement on the data set. In fraud cases, the classification values need to be verified by fraud investigators and the testing rerun with possible changes made to the data set before making the determination on the best model(s) for production.  

The practical solution was a two-fold process for using a test harness that iterates through all possible useful relevant algorithms and sampling sets since each data set has a unique data structure, as well as data scientists working with an interdisciplinary team to improve data accuracy and model prediction using a continual feedback loop through the interrelated departments handling the transactions. New similar studies aggregating research results could yield greater insight into the science of understanding fraud data characteristics. 

\begin{flushleft}
\textbf{Declaration of Competing Interest}\\
The authors declare that they have no known competing financial interests or personal relationships that could have appeared to influence the work reported in this paper. This research did not receive any specific grant from funding agencies in the public, commercial, or not-for-profit sectors.
\end{flushleft}

\printcredits

\bibliographystyle{cas-model2-names}

\bibliography{cas-refs}

\end{document}